\documentclass[]{beingbeyond}
\usepackage{enumitem}
\usepackage[toc,page,header]{appendix}

\usepackage[utf8]{inputenc} 
\usepackage[T1]{fontenc}    
\usepackage{hyperref}       
\usepackage{url}            
\usepackage{array}          
\usepackage{booktabs}       
\usepackage{amsfonts}       
\usepackage{nicefrac}       
\usepackage{microtype}      
\usepackage{xcolor}         
\usepackage{xspace}
\usepackage{bm}
\usepackage{bbding}
\usepackage{bbm}
\usepackage{tabularx}
\usepackage{amssymb}
\usepackage{enumitem}
\usepackage{amsmath}
\usepackage{mathtools}
\usepackage{amsthm}
\usepackage{multirow}
\usepackage{makecell}
\usepackage{color}
\usepackage{colortbl}
\usepackage{adjustbox}
\usepackage{caption}
\usepackage{graphicx}
\usepackage{soul} 
\usepackage{pifont}
\usepackage{wrapfig}
\usepackage{multicol}

\usepackage{xspace}
\makeatletter
\DeclareRobustCommand\onedot{\futurelet\@let@token\@onedot}
\def\@onedot{\ifx\@let@token.\else.\null\fi\xspace}

\def\ie{\emph{i.e}\onedot}

\definecolor{BlockC}{gray}{0.98}  
\definecolor{BlockA}{RGB}{191,211,230}
\definecolor{BlockB}{RGB}{199,233,192}

\definecolor{cgreen}{RGB}{0, 150, 0}
\definecolor{cred}{RGB}{200, 0, 0}
\definecolor{cblue}{RGB}{0, 0, 150}

\title{Being-H0.7: A Latent World-Action Model\\ from Egocentric Videos}

\author{
\textbf{BeingBeyond Team}
}

\webpage{\url{https://research.beingbeyond.com/being-h07}}

\firstfig[width=\linewidth][\textwidth]
  {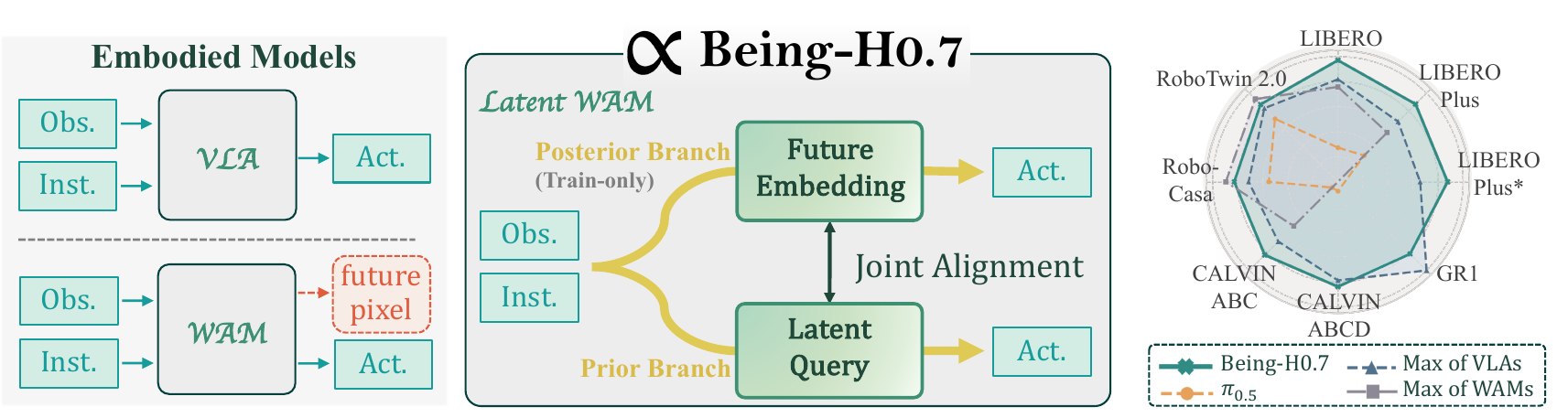}
{
\textbf{Being-H0.7 at a glance.} We build a Latent World-Action Model that differs from VLAs and WAMs. A latent reasoning space is introduced via a set of latent queries in the prior branch, and is further endowed with world modeling by the joint alignment with a future-aware posterior branch. Pretrained on large-scale egocentric videos, Being-H0.7 achieves strong performance across diverse robot tasks.
\vspace{5mm}
}
  {fig:first_fig_label}

\abstract{
Visual-Language-Action models (VLAs) have advanced generalist robot control by mapping multimodal observations and language instructions directly to actions, but sparse action supervision often encourages shortcut mappings rather than representations of dynamics, contact, and task progress. 
Recent world-action models introduce future prediction through video rollouts, yet pixel-space prediction is a costly and indirect substrate for control, as it may model visual details irrelevant to action generation and introduces substantial training or inference overhead. 
We present \textbf{Being-H0.7}, a \emph{latent world-action model} that brings future-aware reasoning into VLA-style policies without generating future frames. 
Being-H0.7 inserts learnable latent queries between perception and action as a compact reasoning interface, and trains them with a future-informed dual-branch design: a deployable prior branch infers latent states from the current context, while a training-only posterior branch replaces the queries with embeddings from future observations. 
Jointly aligning the two branches at the latent reasoning space leads the prior branch to reason future-aware, action-useful structure from current observations alone. 
At inference, Being-H0.7 discards the posterior branch and performs no visual rollout. 
Experiments across six simulation benchmarks and diverse real-world tasks show that Being-H0.7 achieves state-of-the-art or comparable performance, combining the predictive benefits of world models with the efficiency and deployability of direct VLA policies.
}

\checkdata[Date]{Apr 14, 2026}

\begingroup

\endgroup

\begin{document}

\maketitle


\section{Introduction}

Generalist robotic policies are rapidly evolving from task-specific controllers to large embodied models capable of following language instructions, perceiving diverse scenes, and executing long-horizon manipulation across embodiments.
A dominant paradigm is the Vision-Language-Action model (VLA)~\cite{brohan2022rt,zitkovich2023rt,kim2024openvla}, which adapts pretrained vision-language representations to directly map observations and instructions to actions~\cite{black2024pi_0,nvidia2025gr00t,luo2025beingh0,luo2026beingh05}.
Despite strong empirical progress, VLAs face a key bottleneck: observations are dense and semantically rich, while action supervision is sparse and highly correlated with demonstrations.
This imbalance encourages shortcut mappings from visual cues to actions, rather than learning intermediate representations of object dynamics, contact, and task progress.
As a result, these models often perform well in-distribution but struggle when robust control requires anticipating how the world evolves through interaction.

World modeling~\cite{lingbot-world,bruce2024genie,alonso2024diffusion,sun2025worldplay} offer a natural way to address this limitation by enabling robots to reason about how scenes may evolve, rather than merely reacting to the current observation.
Recent world action models (WAM)~\cite{ye2026world,kim2026cosmos,li2026causal,yuan2026fast} introduce future prediction into robot learning, often leveraging video generation models to couple visual rollouts with action generation~\cite{wan2025wan,yang2024cogvideox,agarwal2025cosmos}.
This ``image-then-act'' paradigm is appealing since future frames provide dense supervision from large-scale unlabeled data.

However, explicit future prediction is not necessarily the right substrate for action generation.
Manipulation rarely requires reconstructing future videos with high visual fidelity.
Instead, it requires inferring compact, action-relevant cues, such as contact, object motion, affordances, and task progress, that determine the next action.
Pixel-space prediction is also highly underdetermined: visually distinct futures may imply the same correct action, while visually plausible futures may be irrelevant or misleading for control.
Dense visual prediction can therefore spend capacity on texture, lighting, background, and appearance details that do not improve action prediction.
This mismatch also creates efficiency bottlenecks.
Image-then-act world models inherit the heavy training cost of video generation, which scales poorly when facing internet-scale video pretraining.
At inference time, methods that roll out future frames introduce extra latency and memory cost at every control step, a serious limitation for dynamic tasks such as catching moving objects or interacting with conveyors.
Even approaches that remove test-time rollout remain tied to video-generation training, inheriting its cost, instability, and sensitivity to imperfect pixel predictions.

These observations suggest a different view: future information should shape the policy's internal reasoning, but it need not be reconstructed as pixels.
A robot policy should learn to ask, before generating actions, ``what future-relevant information matters for control?'' rather than ``what will the next image look like?''
This calls for a latent world-action model: a framework that keeps the deployability of direct action prediction while introducing an explicit latent reasoning space where future-aware, action-useful structure can be organized.
Such a space should be compact enough to avoid pixel-level redundancy, expressive enough to capture dynamics and affordances, and tightly coupled to action generation so that future modeling improves control rather than video fidelity.

In this work, we present \textbf{Being-H0.7}, a latent world-action model pretrained from large-scale egocentric videos. 
Instead of predicting future frames, Being-H0.7 introduces a small set of learnable latent queries between the multimodal context and the action tokens.
During model propagation, these dedicated reasoning slots attend to the instruction, observation history, and robot state, aggregate task- and interaction-relevant information, and form a compact latent state before actions are generated.
In this way, action prediction is no longer forced to map directly from raw multimodal context to low-level control.
The model first forms an internal latent representation of what matters for the upcoming interaction, then conditions action generation on this representation.

The key challenge is to train latent queries to support predictive reasoning, even though action-relevant cues are not explicitly annotated.
We address this with a future-informed dual-branch design.
In the deployable \textbf{prior branch}, learnable latent queries are sloted to infer action-useful predictive factors from current multimodal context.
In the \textbf{posterior branch}, these queries are replaced with embeddings extracted from subsequent observations, while the rest of the architecture remains unchanged.
This branch provides an implicit training target: what information, if revealed by later observations, is useful for action prediction?
By aligning the hidden states of the two branches at the latent reasoning positions, the prior queries learn to infer posterior-like reasoning from the current context alone.
Thus, subsequent observations serve as privileged supervision during training, not as a deployment-time requirement.
At inference, the posterior branch is removed entirely: Being-H0.7 performs no future-frame generation or pixel-space rollout before acting. 
It preserves the efficiency of VLAs while injecting a world-modeling signal directly tied to action generation.
To make this latent future alignment stable and scalable, we regularize the latent states with norm and rank constraints, preventing magnitude shrinkage and directional collapse. 
We implement the dual-branch design by packing both branches into a single Mixture-of-Transformers sequence with a dual-branch attention mask, which shares context computation while keeping the two reasoning pathways structurally aligned.

Extensive experiments show that this latent route is both effective and deployable. 
In simulation, Being-H0.7 achieves state-of-the-art or comparable performance across six benchmarks.
In the real world, we evaluate Being-H0.7 on three robot platforms across 12 challenging tasks covering dynamic scenes, physical reasoning, motion reasoning, long-horizon execution, and generalization.
Being-H0.7 leads all five ability-oriented suites, including tasks such as catching a fast rolling ball, pouring into a moving container, folding garments, scanning and sorting packages on a conveyor, and hammering a nail. 
Meanwhile, the deployment stack remains efficient: with latency-aware universal asynchronous chunking (UAC), Being-H variants operate in the 3–4 ms/step regime without adding the burden of test-time future generation.

Our contributions are summarized as follows.
First, we revisit world-action modeling for robot control and argue that prediction should operate in an action-oriented latent space rather than pixel space.
This reframes world modeling as learning compact, control-relevant predictive factors, avoiding the training and inference inefficiencies of image-then-act pipelines.
Second, we introduce Being-H0.7, a latent world-action model that uses learnable latent queries as an explicit reasoning interface between perception and action. 
A future-informed posterior branch supervises this interface during training, while the deployable prior branch infers the latent predictive state from the current context alone.
Third, we develop an efficient dual-branch formulation with hidden-state alignment and lightweight regularization to prevent latent collapse, enabling scalable pretraining on large-scale egocentric videos. Finally, we show that Being-H0.7 achieves strong performance across diverse simulation benchmarks and real-world tasks, combining the predictive benefits of world models with the efficiency and deployability of direct VLA-style policies.

\section{Related Work}

\textbf{Vision-Language-Action Models.}
Recent advances in robotic manipulation~\cite{berscheid2019robot, dasari2019robonet, fang2023rh20t, shafiullah2023bringing} have shifted from narrow, single-task specialists toward generalist models trained on diverse, large-scale datasets.
Among them, Vision-Language-Action models (VLAs)~\cite{brohan2022rt,zitkovich2023rt,kim2024openvla,janner2022planning,chi2025diffusion,feng2025spatial,wang2026rethinking} adapt pretrained vision-language models (VLMs)~\cite{steiner2024paligemma, li2025eagle, wang2024qwen2vl, zhang2025bpe, zhang2025unified, hao2025openmmego, feng2025videoorion} for robotic control, and are effective at predicting actions directly from current observations.
A key line of progress lies in action-head design: early methods rely on autoregressive tokenized actions~\cite{kim2024openvla, qu2025spatialvla}, while recent approaches increasingly adopt diffusion-based~\cite{ho2020denoising, lipman2022flow} generators~\cite{li2024cogact, janner2022planning, liu2024rdt, liang2025discrete, black2024pi_0, nvidia2025gr00t}, which improve efficiency and precision for complex control~\cite{intelligence2025pi05, wen2025dexvla,zhong2025dexgraspvla}.
To better support high-level reasoning, some works further introduce textual planning or structured intermediate representations, including Chain-of-Thought (CoT~\cite{wei2022chain}) planning~\cite{zawalski2024ecot, lin2025onetwovla, clark2025action} and spatial abstractions~\cite{li2025spatial} such as bounding boxes~\cite{griffin2023mobile}, dense correspondence fields~\cite{laskin2020curl}, 3D points~\cite{ten2017using}, or trajectory traces~\cite{lee2025molmoact}. However, these methods still primarily infer actions from the current observation, without explicitly modeling how the world may evolve through interaction.

\textbf{World-Action Models.}
Recent work has increasingly explored video generation and world modeling as a foundation for robot control, motivated by the observation that video models capture temporal dynamics and plausible future evolution that are largely absent from static vision-language pretraining. One line of work uses video models primarily as predictive representation learners or transferable world priors, followed by separate action decoding modules~\cite{hu2024video,pai2025mimic,feng2025vidar,liao2025genie}. 
A second line moves toward tighter coupling by jointly modeling future video and action within a unified architecture~\cite{li2025unified,zhu2025unified,liang2025video}, showing that jointly predicting visual futures and action sequences can improve generalization and data efficiency. 
More recent works build on increasingly stronger pretrained video foundation
and further push toward unified world-action models that support closed-loop control, causal rollout, or planning over predicted futures~\cite{bi2025motus, shen2025videovla}. DreamZero~\cite{ye2026world}, Cosmos Policy~\cite{kim2026cosmos}, and LingBot-VA~\cite{li2026causal} exemplify this trend, showing that stronger video priors can be carried into embodied policies to improve generalization and embodiment transfer. Fast-WAM~\cite{yuan2026fast} shows that retaining video co-training during training while removing test-time future generation can preserve strong action performance with substantially lower latency. Our work is most closely related to this emerging world-action modeling line, but differs in that we do not rely on explicit future video rollout. Instead, we use future information to jointly shape a \emph{latent reasoning space} that directly participates in action generation.

Additionally, a related but distinct trend explores future prediction or future-aware alignment in latent space for action generation~\cite{luo2025learning,vujinovic2025act,zheng2025flare,sun2026vla}. 
Recent developments along this direction have taken several forms: introducing multiple supervision of future representations~\cite{zhang2025dreamvla,liu2026last,zhao2026frappe,zhang2026conservative}, aligning action-side hidden states with future-queried states~\cite{luo2026jointalignedlatentaction}, and compressing future observations into action-useful conditions~\cite{su2026world}. 
However, the future signal is often either attached to the action prediction pathway or introduced through a staged procedure that first learns a future-derived target and then predicts it from the current context. 
Our method instead places the future-aligned object in explicit latent reasoning queries before action generation, and jointly aligns prior and posterior branches on these queries, enabling the latent space to adaptively organize future-aware and context-inferable information as a form of latent thinking for action decoding.

\textbf{Human-Centric Learning.}
The high cost of collecting large-scale robot demonstrations has motivated human-centric learning, which seeks to transfer interaction priors from human behavior to robot policies. 
One route lowers the data-collection barrier through portable physical interfaces such as UMI~\cite{chi2024umi,xu2025dexumi,ha2024umi-leg}, which have recently been scaled to over 10,000 hours of robot-compatible demonstrations~\cite{genrobot2025realomin10kh}. 
Another route learns directly from abundant human video corpora, whose broad coverage of environments, objects, and long-horizon interactions has been studied from traditional recognition and grounding~\cite{feichtenhofer2019slowfast,lin2023univtg} to large-scale egocentric benchmarks such as Ego4D~\cite{grauman2022ego4d}, Ego-Exo4D~\cite{grauman2024egoexo4d}, EPIC-KITCHENS~\cite{damen2018epic}, and EgoDex~\cite{hoque2025egodex}. 
Early efforts exploited such videos mainly through representation learning~\cite{nair2022r3m,ma2022vip}, while recent works introduce more structured supervision, including latent action representations as intent abstractions~\cite{yang2025learning,chen2025villa,luo2025predictive,ye2024latent,bu2025univla,oquab2023dinov2}, future-state or video prediction as a proxy for dynamics understanding~\cite{wu2023unleashing,cheang2024gr2,bharadhwaj2024gen2act}, and explicit geometric cues such as point trajectories, keypoints, and bounding boxes~\cite{chen2025vidbot,ma2025glover++,gavryushin2025maple,bahl2022human,bahl2023affordances,wen2023atm,yang2025magma,team2025gemini}. 
Moving closer to direct policy supervision, another line recovers human actions from videos via hand pose, wrist motion, or retargeted manipulation trajectories for VLA pretraining~\cite{kareer2025egomimic,feng2025vipa,punamiya2025egobridge,zhu2025emma,li2025scalable,bi2025h,yang2025egovla}, with Being-H0~\cite{luo2025beingh0} scaling this paradigm through motion-tokenized reconstructed human hand trajectories. 
Being-H0.5~\cite{luo2026beingh05} further generalizes this direction toward unified cross-embodiment pretraining. 
Our work follows this data-centric scaling route and introduces latent world-action modeling as a reasoning form to inject future-aware structure into human-robot VLA pretraining.

\section{Method}

\subsection{Latent Reasoning: At the Crossroads of VLA and World-Action Model}
\label{sec:latent_reason}

\begin{figure}
    \centering
    \includegraphics[width=1\linewidth]{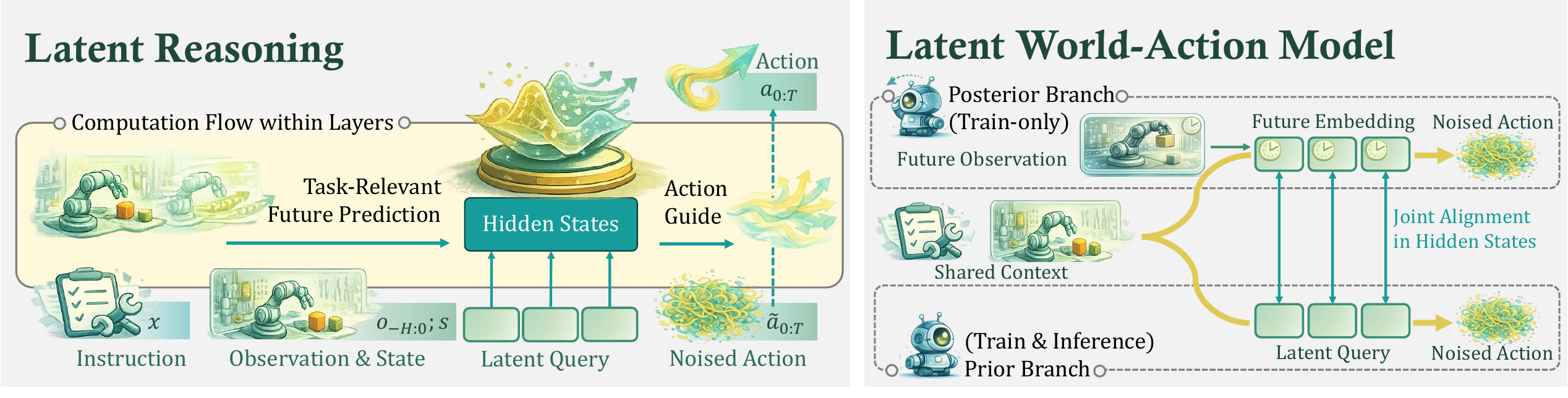}
    \caption{
\textbf{Latent reasoning and latent world-action model.}
\textbf{Left:} Learnable latent queries are inserted to form a latent reasoning space that progressively organizes intermediate hidden states and guides action generation through propagation.
\textbf{Right:} Through joint alignment between the dual-branch design, the model learns to reason with future information at inference time, turning into a latent world-action model.
}
    \label{fig:intuition}
\end{figure}

An effective embodied model should not only react to the instant context but also truly understand how the interaction may unfold. The progress in VLAs and video-generative world-action models highlights these two complementary aspects. Standard VLA models excel at directly mapping current observations to actions, but they do not explicitly model how the world may evolve under interaction. In contrast, video-generative world models attempt to capture such future evolution through dense pixel prediction, but this is both computationally expensive and poorly matched to the abstraction of physical dynamics. We argue that the key is not to choose between these two paradigms, but to connect them through a latent reasoning space: an explicit intermediate space where future-relevant, action-oriented information can be organized before generating low-level actions.

As illustrated in Fig.~\ref{fig:intuition} (Left), we instantiate this idea by introducing a small set of learnable latent queries into the backbone, placing them between the multimodal context and the noised actions. Concretely, let $x$ denote the instruction, $o_{-H:0}$ the observation context of horizon $H$, and $s$ the state. We insert a set of latent queries $Q \in \mathbb{R}^{K \times d}$ before the action chunk, yielding the augmented sequence
\begin{equation}
S = \big[x;\, o_{-H:0};\, s;\, Q;\, a_{0:T}\big],
\end{equation}
where $K$ is the number of latent queries, $d$ is the hidden dimension, and $a_{0:T}$ denotes an action chunk of length $T$. These latent queries define the latent reasoning space, which participates in the layer-by-layer Transformer propagation together with the instruction, observation, state, and action. Through repeated interaction across layers, they progressively integrate task-relevant information from the multimodal context, organize it into an action-oriented latent state, and in turn shape downstream action generation. In this way, the model is no longer forced to map abstract multimodal semantics directly into dense low-level actions. Instead, it can gradually form a compact intermediate reasoning state during forward propagation and use it to guide action prediction.

However, this formulation alone does not guarantee that future prediction will actually emerge within the latent reasoning process. When trained only with action supervision, the latent may instead collapse to a weak intermediate representation or encode only shallow cues sufficient for local action decoding. We therefore introduce in the next subsection a future-informed alignment mechanism to explicitly shape this latent reasoning as world modeling.

\subsection{Latent World-Action Model: Joint Alignment with Future Information}

While the latent reasoning space introduced in Sec.~\ref {sec:latent_reason} provides an explicit substrate for intermediate reasoning, it does not by itself guarantee that the latent queries will organize meaningful future-relevant structure. To shape this latent reasoning space with future information while preserving a deployable inference pathway, we introduce a dual-branch training design, as illustrated in Fig.~\ref{fig:intuition} (Right).

\paragraph{Dual-Branch Design.} We construct two structurally matched branches that share the same context, backbone, and action generation pathway. The \emph{prior branch} is the main deployable branch, where the action generation 
is conditioned only on the current instruction, observation context, state, and a set of learnable latent queries. In parallel, we introduce a training-only \emph{posterior branch}, which has access to the future observations $\tilde{o}_{0:T}$. We replace the latent queries in the posterior branch with a compact set of future embeddings of the same shape, so that the two branches remain structurally aligned at the latent reasoning positions. Concretely, the future observations are first encoded by a frozen pretrained ViT, and then aggregated by a Perceiver resampler into $K$ future embeddings,
\begin{equation}
z^{\mathrm{post}} = E(\tilde{o}_{0:T}) \in \mathbb{R}^{K \times d},
\end{equation}
where $E$ denotes the temporal visual encoder composed of the frozen ViT and the Perceiver resampler. Here, $K$ matches the number of latent queries in the prior branch, and $d$ is the hidden dimension. 
Under action supervision, the two branches naturally capture different views of reasoning for action generation. The prior branch encourages the model to first organize a latent reasoning state from the current context and then generate actions from this latent reasoning state. In contrast, the posterior branch is to reveal which future information is truly useful for action decision-making. By replacing the latent queries with future embeddings, it provides a future-informed version of the reasoning space and highlights the part of future evolution that should matter for downstream action generation.

\paragraph{Joint Alignment.} We then introduce joint alignment on the hidden states of the two branches at the latent reasoning positions, so that these two views explicitly meet in the same latent space. Formally, let $h_\ell^{\mathrm{prior}}$ and $h_\ell^{\mathrm{post}}$ denote the matched latent hidden states at the $\ell$-th aligned layer from the prior and posterior branches, respectively. 
We apply the following point-wise alignment loss:
\begin{equation}
\mathcal{L}_{\mathrm{align}}
=
\frac{1}{L}
\sum_{\ell=1}^{L}
\frac{1}{|h_\ell|}
\left\|
h_{\ell}^{\mathrm{prior}} - h_{\ell}^{\mathrm{post}}
\right\|_F^2,
\end{equation}
where $L$ is the number of aligned layers, $\|\cdot\|_F$ denotes the Frobenius norm, and $|h_\ell|$ denotes the number of scalar elements in the matched latent hidden states at layer $\ell$.
Through this future-informed joint alignment, the latent reasoning space is no longer merely an intermediate carrier for action decoding. Instead, it is explicitly shaped to encode future-relevant, action-oriented structure. In this sense, the resulting model can be viewed as a \emph{latent world-action model}: future information is introduced only during training, yet its effect is realized through the latent reasoning pathway that remains fully executable at inference time.

\subsection{Efficient Dual-Branch Implementation}

\begin{figure}
    \centering
    \includegraphics[width=0.95\linewidth]{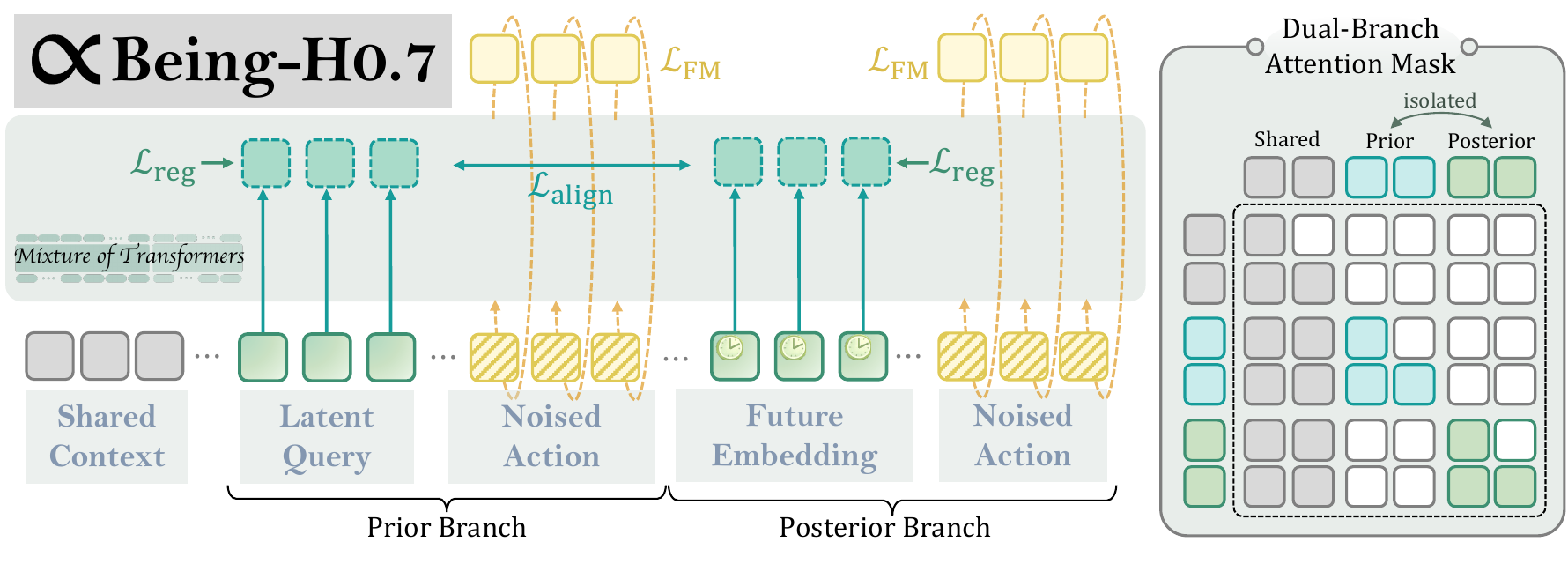}
    \caption{ \textbf{Being-H0.7 Architecture.} We pack the prior and posterior branches into a single MoT sequence with shared context, where the two branches are optimized simultaneously. The posterior branch replaces latent queries with future embeddings, and the two branches are coupled by hidden-state alignment and lightweight regularization. A dual-branch attention mask is applied to isolate prior and posterior branches while preserving access to the shared context for efficient training.
}
    \label{fig:arch}
\end{figure}

We implement the latent world-action model in a structurally-simple and training-efficient way, as illustrated in Fig.~\ref{fig:arch}. We adapt a Mixture-of-Transformers (MoT)~\cite{liang2025mixtureoftransformers} structure like Being-H0.5~\cite{luo2026beingh05}, where action and state vectors are processed with a specific Action Expert and other signals are processed by a larger Understanding Expert. Instead of running two fully separate forward passes, we pack the prior and posterior branches into a single sequence. The two branches share the same current context tokens, while their branch-specific tokens occupy different latent reasoning positions: the prior branch uses learnable latent queries before actions, and the posterior branch uses future embeddings of the same shape.

To preserve the intended dual-branch structure within one packed sequence, we apply a dual-branch attention mask. 
Shared context tokens are visible to both branches, while the prior and posterior branch tokens are not allowed to attend to each other. Thus, the two branches are coupled only through the alignment loss applied to corresponding latent reasoning positions, rather than through direct cross-branch attention. 
In addition, we assign identical positional IDs to corresponding prior and posterior token positions, ensuring that latent queries and future embeddings remain structurally matched throughout the Transformer layers.  This design enables the model to efficiently realize the latent world-action modeling through the dual-branch formulation within a single backbone forward pass.

For action generation, we apply a flow-matching objective to both prior and posterior branches. 
Let $a$ denote the ground-truth action chunk, $t\in[0,1]$ the sampled flow time, and $\epsilon\sim\mathcal{N}(0,I)$ the Gaussian noise. 
We construct the interpolated action $a_t = t a + (1-t)\epsilon$ and target velocity $u_t = a - \epsilon$ along the linear probability path. 
Let $c=[x;\,o_{-H:0};\,s]$ be the shared current context, $q$ the learnable latent reasoning queries, and $z^{\mathrm{post}}$ the future embeddings used by the posterior branch. 
The two branches predict velocity fields $v_\theta^{\mathrm{prior}}(a_t,c,q)$ and $v_\theta^{\mathrm{post}}(a_t,c,z^{\mathrm{post}})$, respectively. 
The combined flow-matching loss is
\begin{equation}
\mathcal{L}_{\mathrm{FM}}
=
\mathcal{L}_{\mathrm{FM}}^{\mathrm{prior}}
+
\mathcal{L}_{\mathrm{FM}}^{\mathrm{post}},
\quad \text{where}  \ 
\mathcal{L}_{\mathrm{FM}}^{\mathrm{prior}}
=
\left\|
v_\theta^{\mathrm{prior}}(a_t,c,q) - u_t
\right\|_2^2,
\quad
\mathcal{L}_{\mathrm{FM}}^{\mathrm{post}}
=
\left\|
v_\theta^{\mathrm{post}}(a_t,c,z^{\mathrm{post}}) - u_t
\right\|_2^2.
\end{equation}

Since hidden-state alignment alone may admit trivial solutions, we apply lightweight anti-collapse regularization to the latent reasoning states of both branches from two aspects: norm preservation and spectral diversity.
For a latent hidden state $h$, we use the norm regularizer
\begin{equation}
\mathcal{R}_{\mathrm{norm}}(h)
=
\left[\mathrm{ReLU}(\tau-\|h\|_2)\right]^2,
\end{equation}
where $\tau$ is a predefined threshold. 
This prevents the aligned states from collapsing toward vanishing magnitude.
For spectral diversity, let $H\in\mathbb{R}^{M\times n}$ denote the projection of a collection of latent hidden states from one branch at one aligned layer onto a random $n$-dimensional subspace, where $M$ is the number of collected latent states. 
We normalize each row of $H$ to unit norm to obtain $\hat{H}$, compute the Gram matrix $G=\hat{H}\hat{H}^{\top}$, and let $\{\lambda_i\}_{i=1}^{M}$ be its eigenvalues. 
With the normalized spectrum $p_i=\lambda_i/\sum_j\lambda_j$, we define
\begin{equation}
\mathcal{R}_{\mathrm{rank}}(H)
=
\sum_{i=1}^{M} p_i \log p_i.
\end{equation}
Minimizing this negative spectral entropy encourages a flatter spectrum and discourages directional collapse of the latent reasoning space. 
Applying both to the aligned latent states across all aligned layers, we obtain
\begin{equation}
\mathcal{L}_{\mathrm{reg}}
=
w_{\mathrm{norm}} \mathcal{R}_{\mathrm{norm}}
+
w_{\mathrm{rank}} \mathcal{R}_{\mathrm{rank}}.
\end{equation}
In practice, we pretrain the model on mixed human and robot manipulation data following the unified sequence format of UniHand~2.0~\cite{luo2026beingh05}. 
This provides a shared interface for cross-embodiment action learning and allows heterogeneous manipulation trajectories to be trained within the same latent world-action framework. 
Although the proposed architecture is compatible with text-generation tasks under the same data format, we focus on action generation in the current stage. 
Together with the prior--posterior alignment loss $\mathcal{L}_{\mathrm{align}}$ defined in the previous section, the final training objective is
\begin{equation}
\mathcal{L}
=
\mathcal{L}_{\mathrm{FM}}
+
w_{\mathrm{align}} \mathcal{L}_{\mathrm{align}}
+
\mathcal{L}_{\mathrm{reg}}.
\end{equation}

\section{Experiments}

\subsection{Training details}
Across all experiments, the policy relies on RGB-only visual observations. 
Context images are uniformly resized to $224\times224$, whereas future frames used by the posterior branch are resized to $256\times256$. 
Being-H0.7 is built on top of Being-H0.5~\cite{luo2026beingh05}, with InternVL3.5~\cite{wang2025internvl3.5} as the understanding expert and Qwen3~\cite{yang2025qwen3} as the action expert. 
For consistent visual embedding spaces across context and future observations, we adopt V-JEPA2.1~\cite{mur2026v} for both visual encoders, while keeping the context-frame encoder trainable.

During pretraining, we use an observation horizon of $H=4$, an action chunk length of $T=20$, and $K=16$ latent reasoning queries. The posterior branch perceives the same number of future embeddings, yielding one-to-one correspondence with the prior latent queries at the aligned positions. 
We apply prior-posterior latent alignment to the last $L=9$ Transformer layers.
We jointly optimize the prior and posterior branches with the flow-matching action objective, the prior--posterior latent alignment loss, and anti-collapse regularization, with $w_{\mathrm{align}}=10^{-3}$ and $w_{\mathrm{norm}}=w_{\mathrm{rank}}=10^{-4}$. 
Pretraining is performed on mixed human and robot manipulation trajectories following the unified sequence format of UniHand~2.0.

For downstream post-training, we optimize only the action-generation objective and latent alignment loss on task-specific demonstrations, without applying the anti-collapse regularizers. 
We use sequence packing to maintain an effective global batch size of approximately 128 trajectory chunks.

\subsection{Simulation}

\begin{table}[t]
\centering
\caption{Benchmark comparison on multiple embodied manipulation tasks. CALVIN denotes ``ABCD $\to$ D'' and CALVIN$^*$ denotes ``ABC $\to$ D'', LIBERO-plus$^*$ denotes finetuning with LIBERO-plus dataset}
\label{tab:benchmark_results}
\resizebox{\columnwidth}{!}{
\begin{tabular}{lccccccccc}
\toprule
Model & Size & LIBERO & LIBERO-plus & LIBERO-plus$^*$ & RoboCasa-50 & GR1 & CALVIN & CALVIN$^*$ & Robotwin2 \\
\midrule
\rowcolor{BlockA!30}
\multicolumn{10}{l}{\textbf{\# VLA}} \\
\rowcolor{BlockA!30}
$\pi$0~\cite{black2024pi_0}       & 3B & 94.4 & 53.6 & -    & 42.4 & -    & - & 3.92 & 65.9/58.4 \\
\rowcolor{BlockA!30}
$\pi$0-FAST\cite{pertsch2025fast}  & 3B & 85.5 & 61.6 & -    & -   & -    & - & -    & -   \\
\rowcolor{BlockA!30}
X-VLA~\cite{zheng2025xvla}        & 0.9B & -    & -    & -    & -    & -    & 4.43 & -  & 72.9/72.8 \\
\rowcolor{BlockA!30}
UniVLA~\cite{bu2025univla}       & \textcolor{red}{8B} & 95.5 & -    & -    & -    & -    & 4.63 & 4.41 & -     \\
\rowcolor{BlockA!30}
gr00t-N1.6~\cite{nvidia2025gr00t}   & 3B & 93.9 & -    & -    & 36.0 & 47.6 & 4.60 & 4.24 & -  \\
\rowcolor{BlockA!30}
$\pi$0.5~\cite{intelligence2025pi05}     & 3B & 96.9 & 77.4 & -    & 41.4 & -    & 4.06 & 4.13 & 82.7/76.8 \\
\rowcolor{BlockA!30}
starVLA~\cite{community2026starvla}      & 4B & 96.5 & 77.0 & -    &  -   & 48.8 & - & - & 88.2/88.3 \\
\rowcolor{BlockA!30}
MINT-4B~\cite{huang2026mimic}      & 4B & 98.7 & 80.1 & \textbf{\textcolor{blue}{84.1}} & -    & -    & 4.57 & - & - \\  
\rowcolor{BlockA!30}
ABot-M0~\cite{yang2026abot}      & 4B & 98.6 & \textbf{\textcolor{blue}{80.5}} & -    & - & \textbf{58.3}    & - & - & 86.1/85.1 \\  
\rowcolor{BlockA!30}
LingBot-VLA~\cite{wu2026pragmatic}  & 4B & -    & -    & -    & -    & -    & - & - & 86.5/85.3 \\
\rowcolor{BlockA!30}
\rowcolor{BlockA!30}
Being-H0.5~\cite{luo2026beingh05}   & 2B & 98.9 & 78.5 & 83.1 & 53.5 & -    & 4.63 & 4.48 & - \\ 
\midrule
\rowcolor{BlockB!30}
\multicolumn{10}{l}{\textbf{\# World Model}} \\
\rowcolor{BlockB!30}
UWM~\cite{zhu2025unified} & - & 79.0 & -    & -    &  48.2    & - & - & - & - \\
\rowcolor{BlockB!30}
UVA~\cite{li2025unified} & - & - & -    & -    &  50.0    & - & - & - & - \\
\rowcolor{BlockB!30}
VPP~\cite{hu2024video} & 1.5B & - & -    & -    &  -    & - & - & 4.33 & - \\
\rowcolor{BlockB!30}
DreamVLA~\cite{zhang2025dreamvla}     & - & 92.6 & - & - & -    & - & - & 4.44 & - \\
\rowcolor{BlockB!30}
JEPA-VLA~\cite{miao2026jepa}     & - & 96.4 & 25.6 & -    & -    & - & - & - & 73.5/- \\
\rowcolor{BlockB!30}
VLA-JEPA~\cite{sun2026vla}     & - & 96.1 & 79.5 & -    & -    & - & - & - & - \\
\rowcolor{BlockB!30}
LingBot-VA~\cite{li2026causal}   & \textcolor{red}{5B} & 98.5 & -    & -    & -    & - & - & - & \textbf{92.9/91.6} \\
\rowcolor{BlockB!30}
Cosmos-Policy~\cite{kim2026cosmos}& {2B} & 98.5 & -    & -    & \textbf{67.1} & - & - & - & - \\
\rowcolor{BlockB!30}
Fast-WAM~\cite{yuan2026fast}     & \textcolor{red}{6B} & 97.6 & -    & -    &  -    & - & - & - & \textbf{\textcolor{blue}{91.9/91.8}} \\
\midrule
Being-H0.7   & 3B & \textbf{99.2} & \textbf{82.1} & \textbf{84.8} & \textbf{\textcolor{blue}{62.1}} & \textbf{\textcolor{blue}{49.2}} & \textbf{4.67} & \textbf{4.48} & 90.2/89.6 \\
\bottomrule
\end{tabular}
}
\end{table}

\subsubsection{Experimental Setup}

We evaluate the Being-H0.7 model on the following six widely-used simulation benchmarks:

\begin{itemize}
    \item \textbf{LIBERO}~\cite{liu2023libero}: 
    LIBERO is a comprehensive benchmark designed to evaluate knowledge transfer and lifelong learning capabilities in tabletop manipulation. 
    It consists of four distinct task suites (Goal, Object, Spatial, and Long). 
    We follow~\cite{liu2023libero,kim2025fine} and train our model on data from all four suites. 
    For evaluation, we conduct 500 trials per suite and report the average success rate across all suites.

    \item \textbf{RoboCasa}~\cite{nasiriany2024robocasa}:
    RoboCasa provides a large-scale simulation framework focusing on everyday long-horizon household tasks. 
    We evaluate 24 base manipulation tasks within diverse kitchen environments and adopt the challenging few-shot setting, utilizing 50 human demonstrations per task. 
    Evaluation is conducted over 50 trials per task across held-out scenes, specifically testing the model's robustness to unseen object instances and novel kitchen styles.

    \item \textbf{GR1}~\cite{nvidia2025gr00t}: 
    GR1 is a bimanual manipulation benchmark featuring a GR-1 humanoid robot equipped with Fourier dexterous hands. 
    It comprises 24 complex tabletop manipulation tasks that require fine-grained dexterity and coordination. 
    We train our model using 1000 demonstrations per task. 
    Evaluation is performed with 50 trials per task.

    \item \textbf{LIBERO-plus}~\cite{fei2025liberoplus}: 
    LIBERO-plus is explicitly designed to systematically assess policy robustness and zero-shot generalization under a diverse set of controlled environmental perturbations. 
    Following standard practice~\cite{fei2025liberoplus}, 
    we evaluate our model under two distinct training configurations: a baseline trained exclusively on the standard LIBERO dataset, 
    and a variant fine-tuned on the augmented LIBERO-plus dataset.

    \item \textbf{RoboTwin 2.0}~\cite{chen2025robotwin}:
    RoboTwin 2.0 is a comprehensive framework designed to benchmark robust bimanual robotic manipulation. 
    To systematically assess and enhance sim-to-real transfer, the benchmark incorporates structured domain randomization along five axes: 
    table-top clutter, varied lighting conditions, diverse background textures, tabletop height variations, and diverse language instructions. 
    We train our model on 2,500 demonstrations from clean scenes (50 per task) and 25,000 from highly randomized scenes (500 per task). 
    We evaluate the policy under two distinct settings: \textbf{Easy} (clean scenes) and \textbf{Hard} (domain-randomized scenes), with 100 rollouts per task.

    \item \textbf{CALVIN}~\cite{mees2022calvin}: 
    CALVIN is a benchmark that specifically targets multi-task learning and long-horizon manipulation capabilities across four distinct environments (A, B, C, and D). 
    Following the standard evaluation protocol, we assess our model on two data splits: ABCD$\to$D (training across all environments and testing on seen environment D) and ABC$\to$D (testing zero-shot generalization to the unseen environment D). 
    The evaluation is rigorously performed over 1,000 unique instruction sequences, where each sequence requires the agent to execute 5 consecutive instructions. 
    We report the average number of tasks completed per sequence. 
\end{itemize}

\subsubsection{Results}

Across all six simulation benchmarks, Being-H0.7 achieves state-of-the-art overall performance, maintaining the highest average rank as detailed in Table~\ref{tab:benchmark_results}.
On \textbf{LIBERO}, 
Being-H0.7 reaches a \textbf{99.2\%} average success rate, 
with strong performance across all suites. 
On \textbf{RoboCasa}, 
our model achieves an exceptional \textbf{62.1\%} success rate,
demonstrating robust proficiency in executing everyday household tasks within diverse and unseen kitchen environments.
Similarly, on \textbf{GR1}, 
it demonstrates strong proficiency in dexterous humanoid manipulation with a \textbf{49.2\%} average success rate.
When evaluating robustness under environmental perturbations on \textbf{LIBERO-plus}, 
Being-H0.7 secures a \textbf{82.1\%} zero-shot success rate, 
which further improves to \textbf{84.8\%} after fine-tuning on LIBERO-plus, highlighting its resilience against shifted viewpoints, novel textures, and sensor noise.
On \textbf{RoboTwin 2.0}, 
Being-H0.7 demonstrates remarkable robustness in complex bimanual manipulation, 
sustaining an \textbf{89.6\%} success rate under severe visual domain randomization, 
with merely a \textbf{0.6\%} performance drop compared to the clean setting (\textbf{90.2\%}).
Finally, on \textbf{CALVIN}, 
Being-H0.7 proves its capacity for multi-task long-horizon execution and zero-shot environment generalization, 
successfully completing an average of \textbf{4.67} and \textbf{4.48} tasks in a row (out of 5) on the ABCD$\to$D and ABC$\to$D splits, respectively.

\subsection{Real-world Experiments}
\label{sec:real_robot}

\begin{figure}
    \centering
    \includegraphics[width=.8\linewidth]{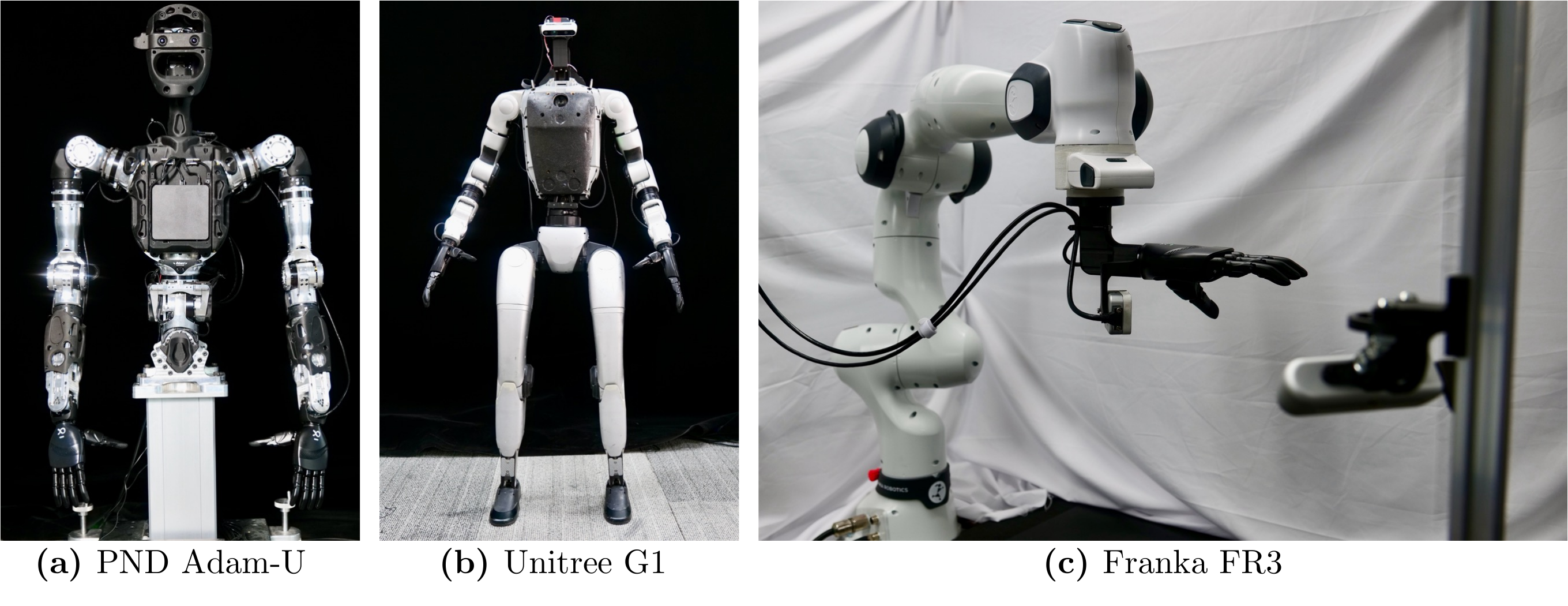}
    \caption{Overview of the real-world embodiments used in this evaluation.}
    \label{fig:exp-embodiments}
\end{figure}

We evaluate Being-H0.7 on three real-robot platforms: \textbf{PND Adam-U}, \textbf{Unitree G1}, and \textbf{Franka FR3}.
All three platforms are equipped with \textbf{Linkerbot O6 (6 DoF)} hands.
PND Adam-U and Unitree G1 use bilateral hand configurations, while Franka FR3 provides a single-arm tabletop setting with one external camera and one wrist camera.
Figure~\ref{fig:exp-embodiments} provides a visual overview of the three deployed embodiments.

\subsubsection{Embodiments and Task Suites}
\begin{table}[t]
\centering
\caption{\textbf{Real-robot embodiments for Being-H0.7.}
All evaluated platforms are paired with Linkerbot O6 (6 DoF) hands and share a unified state/action interface.}
\label{tab:real_robot_platforms_h07}
\small
\setlength{\tabcolsep}{4pt}
\renewcommand{\arraystretch}{1.15}
\resizebox{\columnwidth}{!}{
\begin{tabular}{lcccccc}
\toprule
\textbf{Platform} & \textbf{Type} & \textbf{Body DoF} & \textbf{Hand} & \textbf{Total DoF} & \textbf{Cameras} & \textbf{Policy Freq.} \\
\midrule
PND Adam-U & Upper-body humanoid & 19 & Linkerbot O6 (6 DoF) & 31 & 2 ego-view cameras & 20 Hz \\
Unitree G1 & Bimanual humanoid & 14 & Linkerbot O6 (6 DoF) & 26 & 1 ego-view camera & 10 Hz \\
Franka FR3 & Single-arm tabletop & 7 & Linkerbot O6 (6 DoF) & 13 & 1 external + 1 wrist & 20 Hz \\
\bottomrule
\end{tabular}
\par}
\end{table}

\begin{table*}[t]
\centering
\caption{\textbf{Real-robot task set for Being-H0.7.}
Each task is assigned one primary suite and optional overlap tags; the prompt is the instruction given to the policy during evaluation.}
\label{tab:real_tasks_h07}
\small
\setlength{\tabcolsep}{5pt}
\renewcommand{\arraystretch}{1.15}
\resizebox{\linewidth}{!}{
\begin{tabular}{c l l l p{8.8cm}}
\toprule
\textbf{ID} & \textbf{Platform} & \textbf{Primary Suite} & \textbf{Overlap Tags} & \textbf{Prompt} \\
\midrule
Task01 & Franka FR3 & Dynamic Scene & Motion Reasoning & Catch the fast rolling ball on the table before it leaves the capture area. \\
Task02 & PND Adam-U & Physical Reasoning & Long Horizon & Use the pipette to transfer the liquid from the source container into the target container accurately. \\
Task03 & Unitree G1 & Motion Reasoning & Dynamic Scene, Physical Reasoning & Hit the ball with the racket so that it lands in the marked target area. \\
Task04 & PND Adam-U & Physical Reasoning & Long Horizon & Pour the liquid from the beaker through the funnel into the receiving container. \\
Task05 & Unitree G1 & Dynamic Scene & Motion Reasoning, Physical Reasoning & Pour the objects in the cup into the moving target container. \\
Task06 & Franka FR3 & Dynamic Scene & Motion Reasoning, Long Horizon, Generalization & Pick the cargo from the moving conveyor and place it on the correct shelf level. \\
Task07 & PND Adam-U & Physical Reasoning & Long Horizon, Generalization & Fold the garment neatly into the target folded shape. \\
Task08 & Unitree G1 & Long Horizon & Dynamic Scene, Motion Reasoning, Generalization & Scan the package on the moving conveyor and sort it to the correct destination. \\
Task09 & Unitree G1 & Long Horizon & Physical Reasoning, Generalization & Insert the shoe tree into the shoe and place the prepared shoe onto the conveyor. \\
Task10 & PND Adam-U & Long Horizon & Dynamic Scene, Generalization & Pick the shoe from the conveyor and pack it into the shoebox. \\
Task11 & Franka FR3 & Generalization & Long Horizon, Physical Reasoning & Pick the tabletop object and place it into the correct drawer level. \\
Task12 & Franka FR3 & Physical Reasoning & Motion Reasoning, Long Horizon & Pick up the hammer and drive the nail into the hole. \\
\bottomrule
\end{tabular}
}
\end{table*}

Table~\ref{tab:real_robot_platforms_h07} summarizes the three deployed embodiments.
Unless otherwise specified, all embodiments share the same unified control interface and online inference infrastructure.
In PND Adam-U, the policy controls 19 body DoF together with bilateral Linkerbot O6 hands.
In Unitree G1, the policy exposes a 26-DoF upper-body action interface, \ie, 14 arm joints plus 12 Linkerbot O6 hand joints.
Franka FR3 provides a 7-DoF arm paired with a single Linkerbot O6 hand.

For \textbf{Unitree G1}, the policy still exposes the same 26-DoF action interface used by the rest of our deployment stack.
The additional backend is a pretrained \textbf{AMO} controller~\cite{li2025amo}, used as the balance-aware low-level whole-body module for humanoid execution.
In our integration, AMO owns the 50\,Hz Unitree body-control loop, predicts lower-body and waist commands conditioned on the latest upper-arm targets, and composes the final body command for execution, while the Linkerbot O6 hands remain controlled through the same hand interface as the other embodiments.
This keeps the upper-body policy interface consistent while providing stable whole-body execution on G1.

\begin{figure*}[t]
    \centering
    \includegraphics[width=\linewidth]{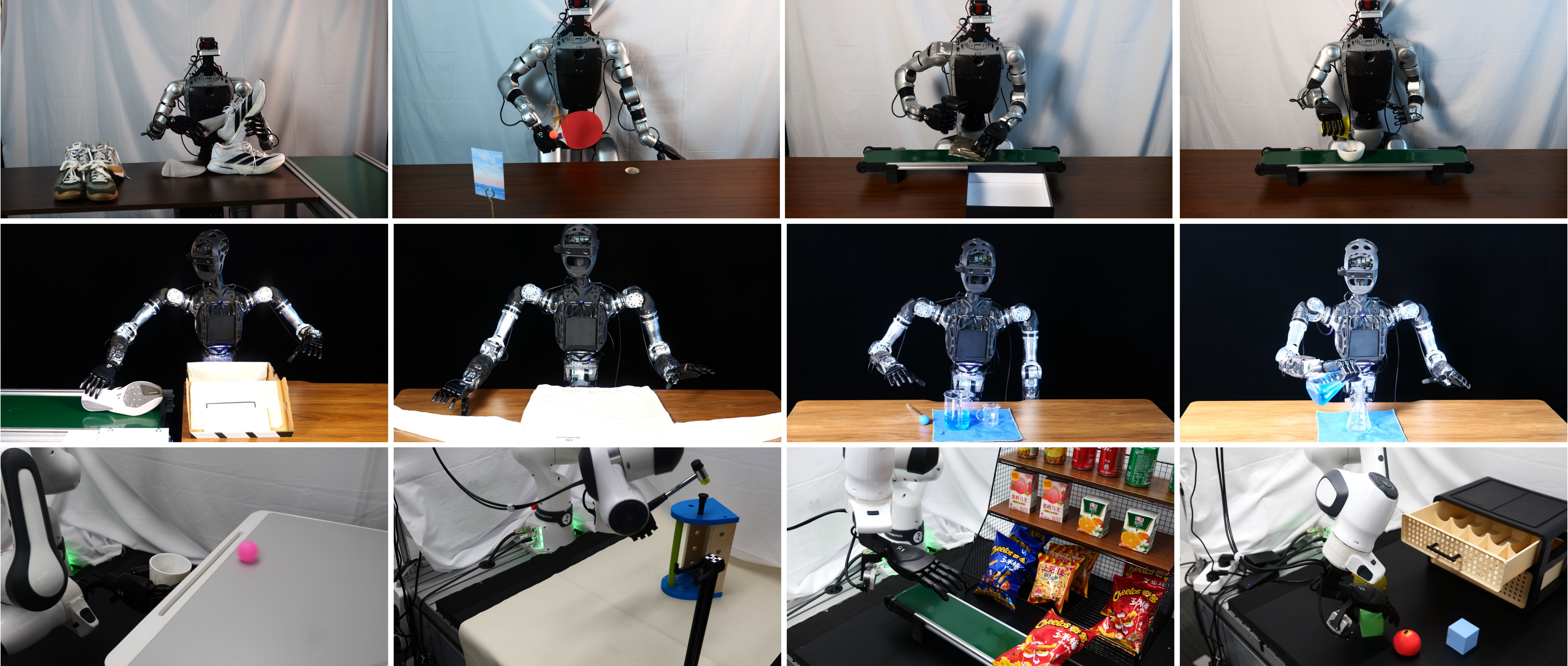}
    \caption{\textbf{Visual overview of the 12 real-robot evaluation tasks.}
    The figure shows the task scenes used in our real-world evaluation across PND Adam-U, Unitree G1, and Franka FR3, covering the five ability-oriented suites.}
    \label{fig:real_exps_h07}
\end{figure*}

We design 12 real-robot tasks for Being-H0.7 and organize them into five \emph{ability-oriented} suites:
\textbf{dynamic scene}, \textbf{physical reasoning}, \textbf{motion reasoning}, \textbf{long-horizon execution}, and \textbf{generalization}.
These suites are intentionally \emph{compositional}: a single task may stress several capabilities at once, such as reacting to moving targets while also reasoning about object trajectories, gravity, containment, or multi-stage subgoals.
Table~\ref{tab:real_tasks_h07} lists the full task set and evaluation prompts.
Figure~\ref{fig:real_exps_h07} provides a unified visual overview of the task scenes used in this evaluation.
For reporting, each task is assigned one primary suite together with optional overlap tags, and suite-level averages are computed over all tasks carrying the corresponding suite tag.

These five suites target distinct difficulty sources.
\textbf{Dynamic Scene} tasks require the policy to react before a moving object or changing scene leaves the feasible interaction window.
\textbf{Physical Reasoning} tasks require predicting consequences induced by gravity, fluid transfer, deformable contact, containment, or tool-mediated interaction.
\textbf{Motion Reasoning} tasks emphasize trajectory anticipation, relative velocity, and contact timing.
\textbf{Long Horizon} tasks stress subgoal memory and sequential consistency across multiple stages.
\textbf{Generalization} focuses on preserving task structure under shifted layouts, shelf levels, containers, and object instances.

\subsubsection{Evaluation Protocol}

We deploy all compared policies through a unified \textbf{black-box inference server}.
This protocol keeps the surrounding execution stack identical across methods.
For each task, we pre-define a set of scene layouts and initial conditions, then randomize both the tested policy endpoint and the rollout order during evaluation.
The operator records task success using a fixed binary criterion defined for that task while the active policy endpoint remains hidden.
Unless otherwise stated, each task is evaluated over \textbf{20 blind trials per method}.

This protocol is especially useful here because several of the new suites explicitly probe \emph{reaction quality} in addition to terminal grasp precision.
For example, tasks involving dynamic scene changes, flexible objects, or liquid-like interaction can be sensitive to small timing differences in policy updates.
The shared deployment server and fixed blind-evaluation procedure keep those comparisons aligned across methods.

\subsubsection{Results Overview}

\begin{figure}[t]
    \centering
    \includegraphics[width=0.8\linewidth]{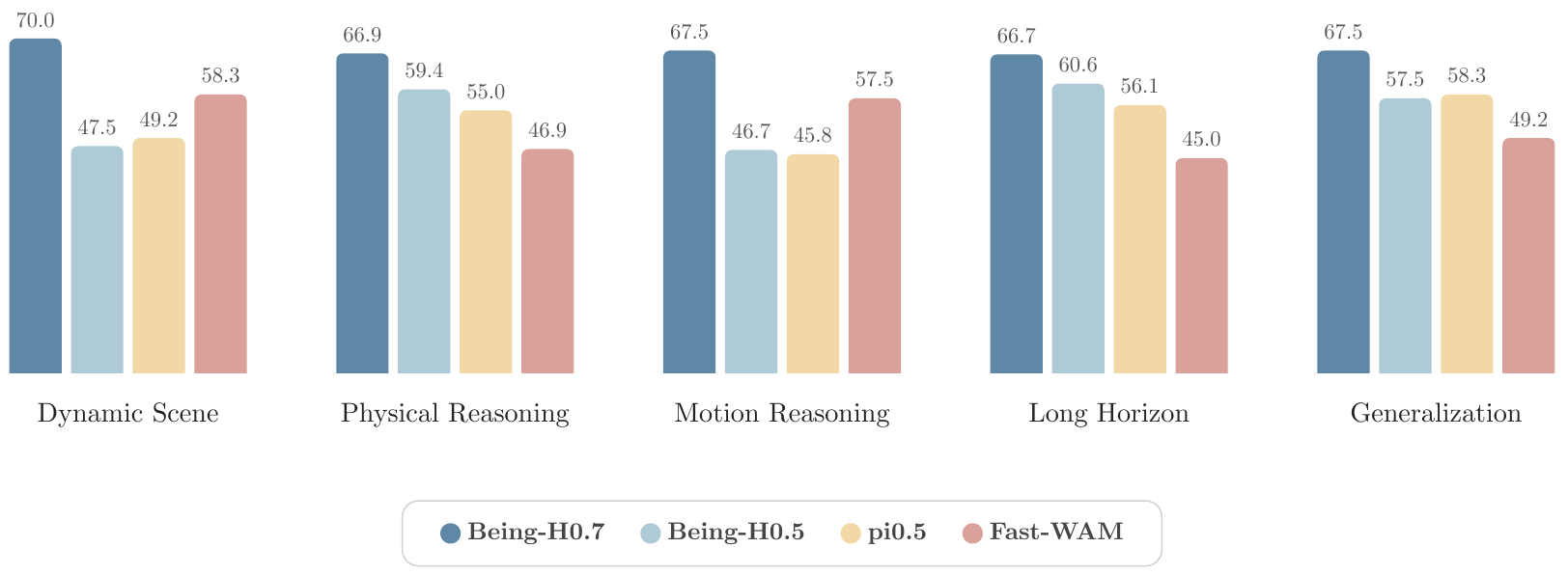}
    \caption{\textbf{Suite-level real-robot success rates (\%).}
    Comparison of Being-H0.7, Being-H0.5, $\pi$0.5, and Fast-WAM on the five ability-oriented task suites.
    Each task is evaluated over 20 blind trials, and each suite score is averaged over all tasks carrying the corresponding suite tag.}
    \label{fig:real_suite_bars_h07}
\end{figure}

Figure~\ref{fig:real_suite_bars_h07} summarizes suite-level success rates
computed from the overlap-tag aggregation described above.
Because the five suites contain different numbers of tagged tasks, the resulting bars are reported with one decimal place.

\textbf{Being-H0.7} leads on all five suites, spanning reactive, physical, sequential, and generalization-oriented tasks across all three embodiments.
This breadth is the main real-robot result of the section: the improvement appears throughout the benchmark rather than concentrating in one corner of it.

\paragraph{Dynamic and motion-centric tasks are where the predictive advantage is most visible.}
The clearest margin appears on \textbf{Dynamic Scene}, and the same ordering largely carries over to \textbf{Motion Reasoning}.
These suites contain the most timing-sensitive tasks in the benchmark, including catching a fast rolling ball, racket-based redirection, pouring into a moving receptacle, and conveyor-based interaction.
Such tasks punish stale state estimates very quickly: once the object has moved beyond the valid contact window, small pose errors or delayed corrections usually lead to immediate failure.
Among the baselines, \textbf{Fast-WAM} remains the strongest one in these reactive suites, which is consistent with its emphasis on low-latency action generation.
Being-H0.7 extends this advantage further, indicating that reactive success is shaped jointly by runtime responsiveness and by a future-aware latent state that tracks object motion, relative timing, and the downstream consequence of committing to a contact.

\paragraph{Physical and long-horizon suites highlight a second strength of the model.}
On \textbf{Physical Reasoning} and \textbf{Long Horizon}, the closest baseline is \textbf{Being-H0.5}, reflecting stronger stable manipulation priors and stage-by-stage execution consistency.
Representative tasks here include pipette transfer, funnel pouring, garment folding, shoe-tree insertion, shoe boxing, and hammer-and-nail interaction.
Success couples dexterity with reasoning about containment, gravity, deformable contact, tool-mediated force transfer, and how earlier subgoals constrain later ones.
Being-H0.7 stays ahead on both suites, showing that the learned world-action prior supports fast reaction and also maintains causal consistency through longer and more physically constrained manipulation chains.

\begin{figure}[ht]
    \centering
    \includegraphics[width=1\linewidth]{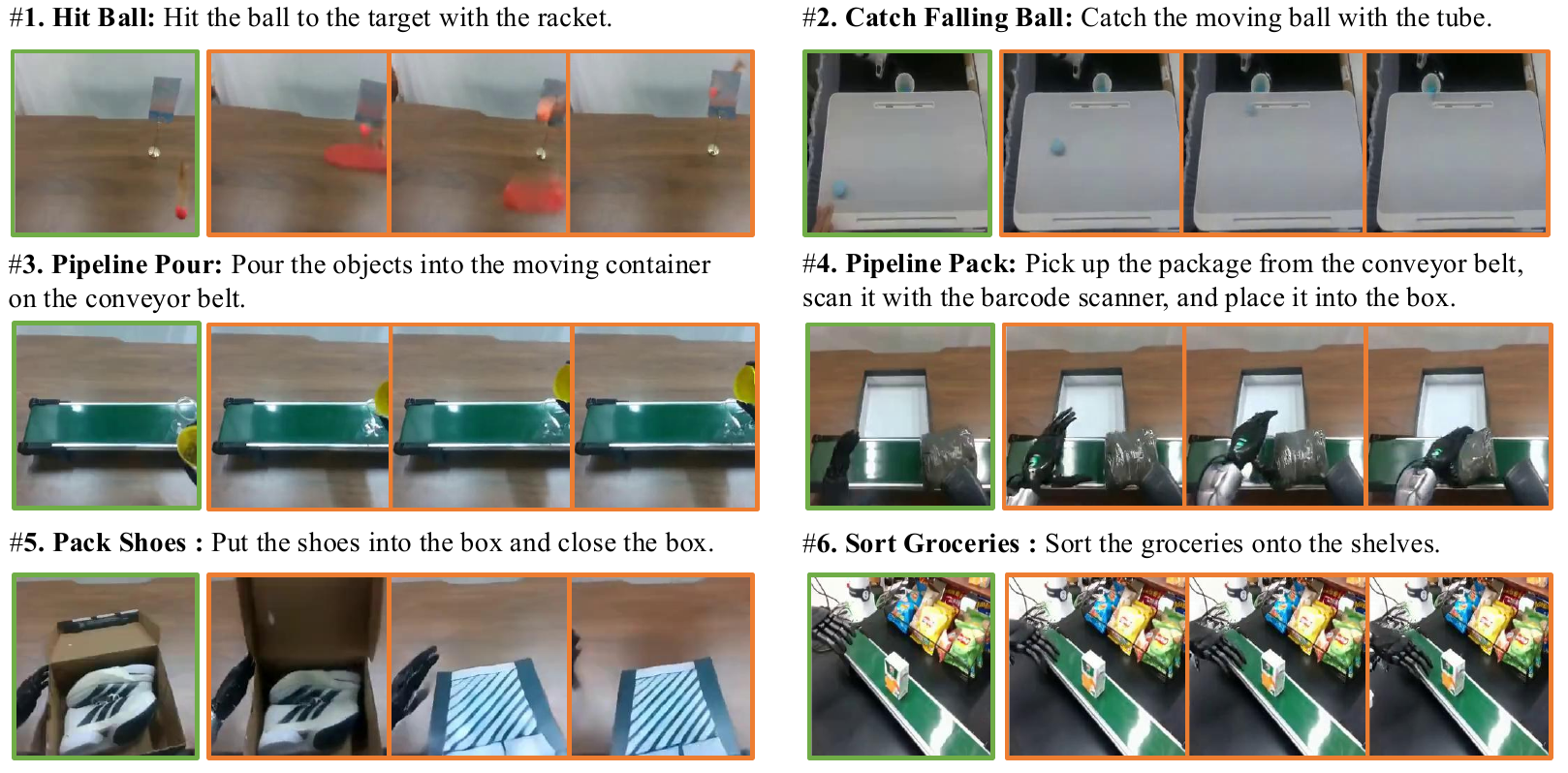}
    \caption{Visualization of the Latent Reasoning.}
    \label{fig:vis_latent}
\end{figure}

\begin{figure}[ht]
    \centering
    \includegraphics[width=0.75\linewidth]{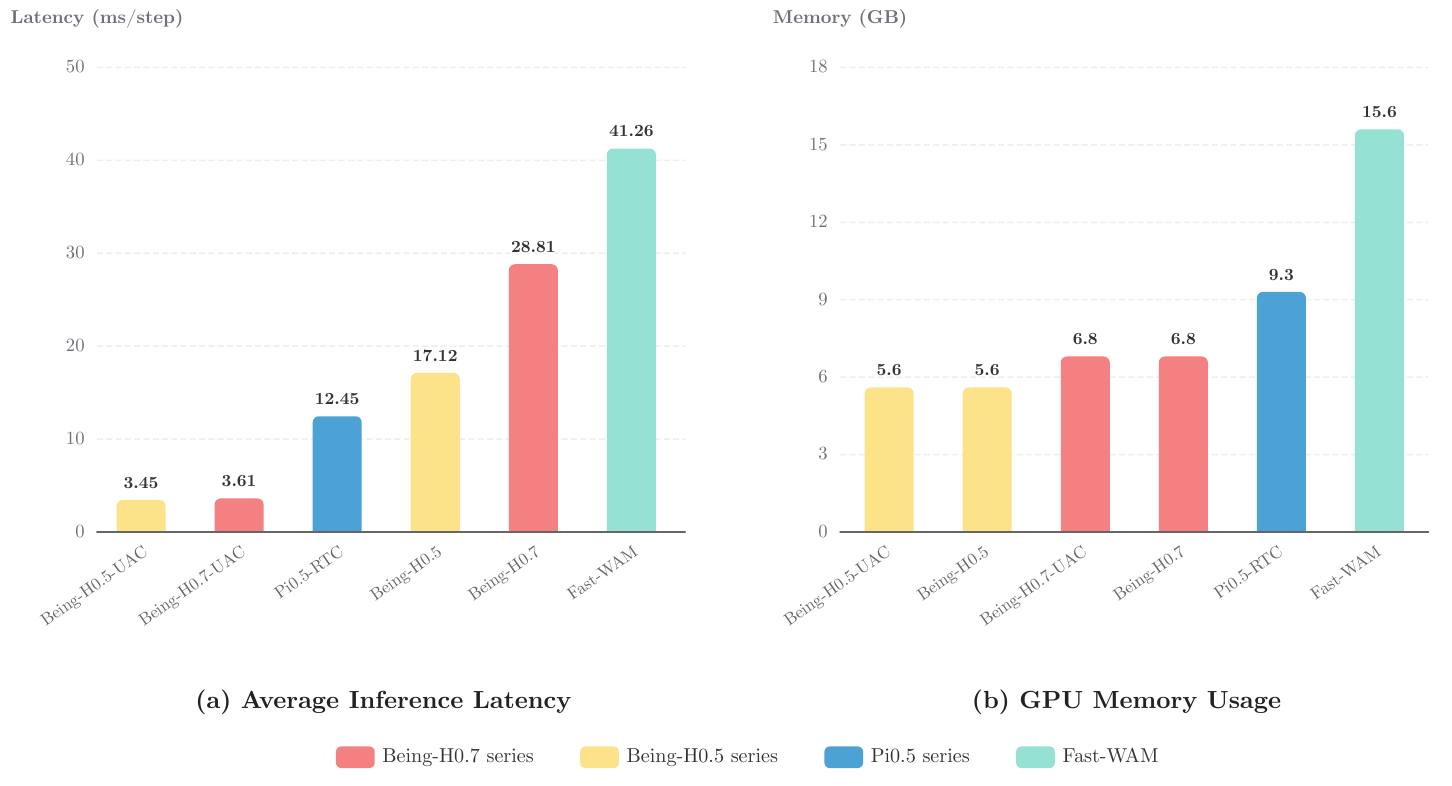}
    \caption{Inference cost measured in the real-world deployment stack. We report it as a system-level view of deployability alongside the main success-rate comparison.}
    \label{fig:inference_cost}
\end{figure}

\paragraph{Generalization improvements persist across embodiments.}
The \textbf{Generalization} suite mixes tasks from all three platforms and stresses shifted layouts, shelf heights, containers, object instances, and camera geometries.
Here, \textbf{$\pi$0.5} and \textbf{Being-H0.5} remain reasonably competitive, making the suite balanced and practically relevant.
Even so, Being-H0.7 remains the most reliable overall, indicating that the benefit of its latent predictive prior carries across changes in embodiment and scene structure.
Because the benchmark is intentionally compositional, a consistent lead across all five suite bars is stronger evidence than an isolated win on a single hand-picked task.

\paragraph{Visualization of the Latent Reasoning}
To more directly reveal what is encoded in this \textit{latent reasoning space}, the current observation and the latent hidden states from the prior branch of Being-H0.7 can be used jointly as conditions for a video generation model to synthesize future task states. 

Although Being-H0.7 does not explicitly reconstruct future frames during inference, the resulting visualizations in Figure~\ref{fig:vis_latent} suggest that its latent representations already capture predictive information about how the world will evolve. This provides further evidence that Being-H0.7 functions as a latent world action model: rather than modeling the future at the pixel level, it learns an internal predictive representation that is sufficient to support action generation.

\paragraph{Inference infrastructure strengthens real-world deployability.}
Figure~\ref{fig:inference_cost} reports system-level inference cost under the same deployment infrastructure.
On the client side, we further employ the latency-aware \textbf{Universal Async Chunking (UAC)} mechanism from Being-H0.5~\cite{luo2026beingh05}, implemented as asynchronous real-time chunking.
Concretely, the client maintains a thread-safe action buffer together with a running estimate of how many control steps will be consumed before the next chunk becomes available.
A control thread pops actions from the committed prefix at the robot frequency, while a parallel inference thread wakes up when the remaining buffer falls below a trigger threshold, fetches the latest observations, and requests the next chunk from the server.
The crucial rule is that UAC never rewrites the already committed prefix: it only stitches the future suffix back into the buffer after the estimated inference delay.
This \emph{prefix-lock / suffix-update} design absorbs model, transport, and scheduling jitter without changing the policy interface itself, and it makes the same deployment protocol usable across platforms with different control frequencies and embodiment-specific action dimensions.
UAC is the deployment protocol that turns chunked prediction into continuous control.
It preserves temporal continuity, reduces visible control stutter, and keeps the evaluation stack uniform across embodiments.
The most visible effect is that the UAC-enabled Being-H variants move into the 3--4\,ms/step regime while keeping the same GPU memory footprint as their non-UAC counterparts.
This gives the controller more timing slack on dynamic tasks, keeps buffer occupancy steadier under network and scheduler jitter, and makes the online rollout feel substantially smoother at the robot interface.
Together with the suite-level results above, the cost plot shows that the policy gains are realized inside an efficient inference loop rather than at the expense of an unwieldy deployment setup.

\section{Conclusion}
We introduced Being-H0.7, a \emph{latent world-action model} that bridges direct action prediction and world modeling through a compact latent reasoning space.
By aligning a deployable prior branch with a future-aware posterior branch, our method injects future-relevant reasoning into action generation without requiring costly pixel-level rollout at inference time.
Combined with large-scale human video pretraining, Being-H0.7 provides an effective and scalable framework for embodied models.

\clearpage

\bibliographystyle{unsrt}
\bibliography{ref}

\clearpage

\beginappendix
\section*{Author List}

Hao Luo$^*$, Wanpeng Zhang$^*$, Yicheng Feng$^*$, Sipeng Zheng$^{*}$, Haiweng Xu, Chaoyi Xu, Ziheng Xi, Yuhui Fu, Zongqing Lu$^{\dagger}$

$^*$Equal Contribution \quad $^{\dagger}$Corresponding Author



\end{document}